\documentclass{article} 
\usepackage[final]{colm2026_conference}

\usepackage{microtype}
\usepackage{amsmath}
\usepackage{graphicx}
\usepackage{subcaption}
\usepackage{hyperref}
\usepackage{url}
\usepackage{booktabs}
\usepackage{enumitem}
\usepackage{lineno}
\usepackage{xcolor}

\usepackage{tcolorbox}
\newtcolorbox{mybox}{colback=gray!5!white,colframe=gray!75!black}

\newtcolorbox{myboxblue}{colback=blue!5!white,colframe=blue!75!black}

\newtcolorbox{myboxmagenta}{colback=magenta!5!white,colframe=magenta!75!black}

\newtcolorbox{myboxolive}{colback=olive!5!white,colframe=olive!75!black}

\definecolor{darkblue}{rgb}{0, 0, 0.5}
\hypersetup{colorlinks=true, citecolor=darkblue, linkcolor=darkblue, urlcolor=darkblue}

\title{Beyond the Assistant Turn: User Turn Generation as a Probe of Interaction Awareness in Language Models}

\author{Sarath Shekkizhar\thanks{Correspondence to: \texttt{sshekkizhar@salesforce.com}} , Romain Cosentino, Adam Earle
}

\begin{document}

\ifcolmsubmission
\linenumbers
\fi

\maketitle

\begin{abstract}
Standard LLM benchmarks evaluate the assistant turn: the model generates a response to an input, a verifier scores correctness, and the analysis ends. This paradigm leaves unmeasured whether the LLM encodes any awareness of what follows the assistant response. We propose \emph{user-turn generation} as a probe of this gap: given a conversation context of user query and assistant response, we let a model generate under the user role. If the model's weights encode \emph{interaction awareness}, the generated user turn will be a grounded follow-up that reacts to the preceding context.
Through experiments across $11$ open-weight LLMs (Qwen3.5, gpt-oss, GLM) and $5$ datasets (math reasoning, instruction following, conversation), we show that interaction awareness is decoupled from task accuracy. In particular, within the Qwen3.5 family, GSM8K accuracy scales from $41\%$ ($0.8$B) to $96.8\%$ ($397$B-A$17$B), yet genuine follow-up rates under deterministic generation remain near zero. In contrast, higher temperature sampling reveals interaction awareness is latent with follow up rates reaching $22\%$. Controlled perturbations validate that the proposed probe measures a real property of the model, and collaboration-oriented post-training on \texttt{Qwen3.5-2B} demonstrates an increase in follow-up rates. Our results show that user-turn generation captures a dimension of LLM behavior, interaction awareness, that is unexplored and invisible with current assistant-only benchmarks.
\end{abstract}

\begin{figure*}[t]
\centering
\includegraphics[width=0.59\textwidth]{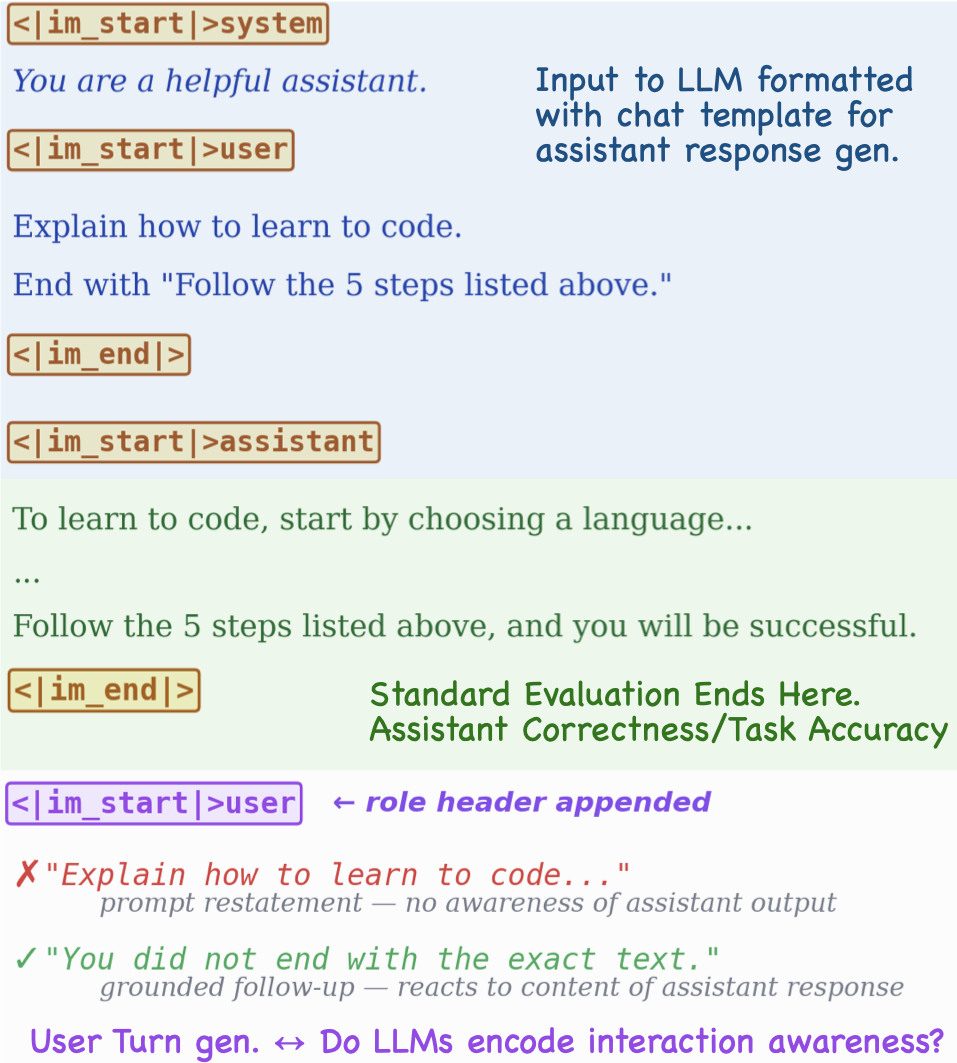}
\hfill
\includegraphics[width=0.39\textwidth]{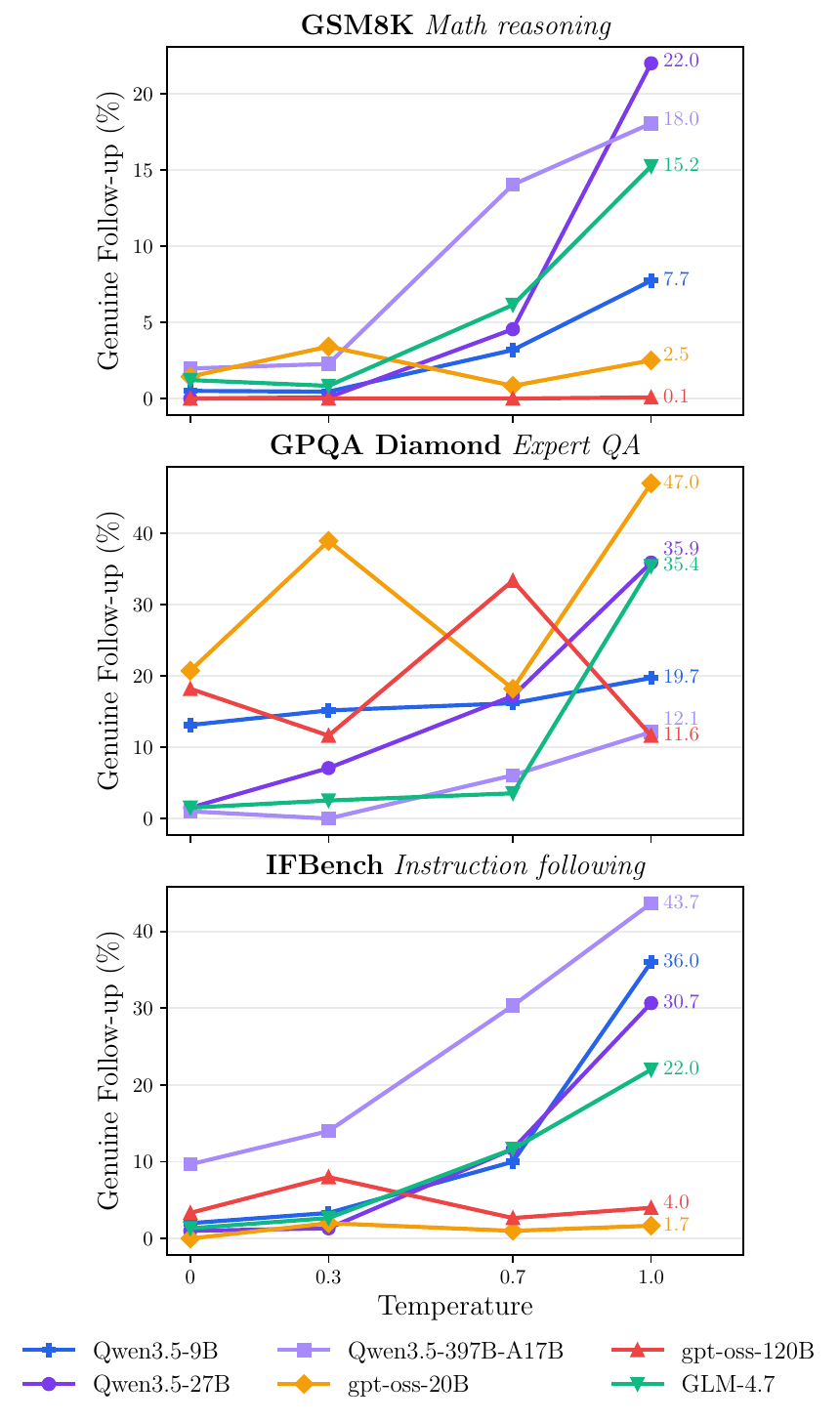}
\caption{\textbf{User-turn generation as a probe of interaction awareness.} \emph{Left:}~A conversation context consisting of system, user, and assistant turns delimited by special tokens (e.g., \texttt{<|im\_start|>}, \texttt{<|im\_end|>}). \emph{Standard evaluation} scores the assistant response for task accuracy. \emph{Our probe} appends a \texttt{<|im\_start|>user} header and lets model~$M_\theta$ generate under the user role. A genuine follow-up indicates the model's weights encode conversational awareness. \emph{Right:}~Genuine follow-up rate (\%) vs.\ sampling temperature for six models across three datasets. At greedy decoding ($T{=}0$), most models produce near-zero follow-ups. Increasing temperature surfaces latent interaction awareness in Qwen3.5 and GLM (e.g., Qwen3.5-27B rises from $0\%$ to $22\%$ on GSM8K), while \texttt{gpt-oss} remains near zero. \texttt{Qwen3.5-9B} starts with higher baseline follow-up on GPQA Diamond ($13.1\%$) and rises further with temperature.}
\label{fig:overview}
\end{figure*}

\section{Introduction}
\label{sec:introduction}
Large language models~(LLMs) are post-trained on token sequences that span both sides of a conversation, i.e., the user and assistant \citep{qwen3.5, openai2025gptoss120bgptoss20bmodel, zeng2025glm}. Yet, standard evaluation benchmarks focus only on the response generated under the assistant role \citep{cobbe2021training, rein2024gpqa, pyatkin2025generalizing}. Consequently, a model can score $96\%$ on GSM8K and yet, when made to continue under the user role, completely fall apart. Failures show up as generating text from a previous turn or continuing as the assistant, a case of identity drift \citep{shekkizhar2025echoing}. The assistant response is correct; however, the most likely generation with user role is not reflective of a real conversation.

We study this gap through \emph{user-turn generation} (Figure~\ref{fig:overview}). Given a prompt, a model first generates an assistant answer. That answer is appended to the chat messages, and the model continues generation under the user role. Since the weights remain same, the generated user turn is a behavioral readout of what the model produces under the user role, a probe of whether the model's training has shaped it to anticipate user reactions to preceding context with assistant output. Our focus is not user simulation \citep{naous2026flipping, dou2025simulatorarena, hathidara2026mirrorbench, zhou2026sim2real}.
Instead, we are interested in surfacing a model's interaction awareness from the user perspective.

This capability is crucial for at least three reasons. First, conversational consequences are central to multi-agent collaboration, self-play, and interactive deployments \citep{schatzmann2007agenda, ShiQWY19, ghandeharioun2019approximating}. Second, understanding the theory-of-mind or partner modeling in an LLM paves the way for better interpretability and collaborative applications \citep{sap2022neural, gandhi2023understanding, gu2024simpletom}. Finally, the use of LLMs in multi-turn conversation often requires user reactions as part of the training \citep{wu2025collabllm, wang2026openclaw, buening2026aligning} to enable better and more capable models. Our work asks a complementary question to these works (see App.~\ref{app:related-work} for additional discussion). We provide an evaluation measure of the interaction capabilities of LLMs.

This work makes the following contributions: \textbf{(i)} We formalize user-turn generation as a probe of interaction awareness and jointly measure assistant accuracy,  follow-up quality in LLMs (\S~\ref{sec:methods}). \textbf{(ii)} Across $11$ models spanning $0.8$B to $397$B parameters, we demonstrate interaction awareness is decoupled from task accuracy across model families (\S~\ref{sec:cross-family}). \textbf{(iii)} We show that this ability can be latent in a model's generation distribution, surfaced with higher temperature sampling, yet does not scale with model size even within a single family (\S~\ref{sec:temperature-results},\ref{sec:scaling}). \textbf{(iv)} We provide controlled perturbations and study collaboration-oriented post-training to validate the proposed follow up metric (\S~\ref{sec:perturbation-results},\ref{sec:post-training-results}).

\section{User Turn Generation}
\label{sec:methods}

Effective conversation requires more than producing correct response for a given input. It requires, in addition, anticipating the consequence to the response provided from the other side. 
Recent work on multi-agent and agent to agent systems shows that such \emph{partner models} are necessary and only emerge when the training environment imposes sufficient pressure for them to matter \citep{monwilliams2025partner}. 
To this end, we ask the question: even without explicitly optimizing for user-turn generation, can current LLMs generate user turns that are grounded in the preceding assistant response?

\subsection{Problem Formulation}
\label{sec:formulation}

Consider an LLM parameterized by $\theta$, $M_\theta$, post-trained with a chat template that defines user and assistant role headers. Given a conversation context consisting of a user query $q$ and an assistant response $a$, we define \emph{user-turn generation} as the sequence produced when $M_\theta$ generates under the user role header, conditioned on the preceding context (Figure~\ref{fig:overview}):
\begin{align}
u &= M_\theta([q; a])
\label{eq:user-turn}
\end{align}
The question of study in this work is whether $u$ constitutes a grounded follow-up to $a$, a reaction that could arise from processing the content of the assistant response. A grounded $u$ is evidence that $\theta$ has learned to produce contextually appropriate user continuations; a degenerate $u$ indicates that, at least under the model's modal continuation, the training has not shaped the weights to generate plausible conversational consequences.

This formulation admits two experimental settings. In the \emph{self-generated} setting, $a$ is itself produced by $M_\theta$ (i.e., $a = M_\theta(q)$), so $u$ probes whether the model can generate the user reaction to its own assistant response. In the \emph{held-out} setting, $a$ comes from a real conversation and the model generates $u$ given the $(q, a)$ pair, testing whether it can produce a user follow-up to a provided response. Both settings use the same model and the same role-switching mechanism; the difference is whether the assistant turn is self-generated or given, i.e., whether the user generation is grounded in its own  response vs independent context.


\subsection{Measuring Interaction Awareness}
\label{sec:metric}
We define a \emph{genuine follow-up} as a generated user turn $u$ whose content is grounded in the context $(q, a)$, i.e., $u$ references, reacts to, or builds upon specific content in the conversation. Conversely, a \emph{degenerate} user turn is one that fails this grounding criterion: verbatim restatement of $q$, continuation of the assistant response under the user role, emission of planning artifacts, or generation of another task query. These failure modes indicate that the model defaults to task-solving behavior without producing a contextually grounded user reaction.

We propose to measure interaction awareness through an LLM-based evaluator \citep{zheng2023judging} that analyzes the full conversation trace. Given the query $q, a, u$, the evaluator performs an assessment:
\begin{align}
\textsc{FollowUpEval}(q, a, u) = \{ \text{rationale}, \ell, \sigma \},
\label{eq:followup-eval}
\end{align}
where $\sigma \in \{0, 1\}$ is a binary genuine-followup judgment and $\ell$ is a descriptive label drawn from one of \texttt{previous\_turn\_restate}, \texttt{new\_task\_prompt}, \texttt{assistant\_turn\_restate}, \texttt{malformed\_artifact}, \texttt{meta\_planning}, \texttt{degenerate\_short}, \texttt{plausible\_followup}, or \texttt{other}. We classify $u$ as a genuine follow-up when $\sigma = 1$, and report the \emph{genuine-followup rate} across a dataset as the primary metric of interaction awareness. The \emph{rationale}, an output variable for model to think, and the descriptive label allow for the judge model to reason about the input and produce a more nuanced assessment.


We use an LLM judge over alternatives such as embedding-based classifiers for two reasons. First, interaction awareness requires assessing whether $u$ is contextually grounded in $a$, a judgment that depends on semantic content and pragmatic plausibility. This parallels how \citet{wu2025collabllm, shekkizhar2025echoing} use LLM judges to evaluate conversation-level metrics. Second, the judge generalizes across the datasets in our study without requiring per-domain classifier training. Judge validation against human annotators yields $\kappa = 0.726$ (see Appendix~\ref{app:judge-validation} and \ref{app:manual-review} for more details on judge analysis).

\begin{figure*}[t]
\begin{minipage}[t]{0.48\textwidth}
\begin{mybox}
{\textbf{Truncation} \hfill \textit{GSM8K}}
\begin{customsmall}
\begin{verbatim}
[Assistant]
1. Total eggs laid per day: 16.
2. Eggs eaten for breakfast: 3.
3. Eggs used for muffins: 4.
4. Remaining eggs: 16 - 3 - 4 = 9.
5. Price per egg: $2.
6. Daily earnings: 9 x $2 = $18.
Answer: 18

[User turn] --> meta_planning (nongenuine)

[Assistant -- Truncated]
1. Total eggs laid per day: 16.
...
4. Remaining eggs: 16 - 3 - 4 = 9.
5. Price

[User turn] Complete your answer.
--> plausible_followup (genuine)
\end{verbatim}
\end{customsmall}
\end{mybox}
\end{minipage}
\hfill
\begin{minipage}[t]{0.48\textwidth}
\begin{mybox}
{\textbf{Explicit Question} \hfill \textit{IFEval}}
\begin{customsmall}
\begin{verbatim}
[Assistant]
... So let us praise the eucalyptus tree,
For all the beauty and the use it brings,
A wild and wonderful legacy,
That lifts the soul and makes the spirit sing.

[User turn] --> meta_planning (nongenuine)

[Assistant -- with appended question]
... So let us praise the eucalyptus tree,
For all the beauty and the use it brings,
A wild and wonderful legacy,
That lifts the soul and makes the spirit sing.

What do you think?

[User turn] The poem is great, but I
requested at least 350 words.
--> plausible_followup (genuine)
\end{verbatim}
\end{customsmall}
\end{mybox}
\end{minipage}
\vspace{-.2cm}
\caption{\textbf{Controlled perturbation examples.} \emph{Left:}~Truncation removes the end of the assistant response, prompting the model to produce a reaction to complete the response. \emph{Right:}~Appending a generic question elicits a grounded critique rather than a prompt restatement. Assistant and user turns in the examples are generated by GLM-4.7. These controls demonstrate that interaction awareness can surface in specific contexts.}
\label{fig:perturbation-examples}
\end{figure*}
\subsection{Experimental Setup}
\label{sec:experimental-setup}

We evaluate user-turn generation across $11$ open-weight models, $5$ benchmark datasets, and $2$ held-out conversational datasets. We complement these with controlled perturbations and collaboration-oriented post-training.

\paragraph{Generation settings.} In the \emph{self-generated} setting, given a prompt $q$, we generate $a$ from $M_\theta$ and then $u$ from the same model under the user role (Eq.~\ref{eq:user-turn}). In the \emph{held-out} setting, used for post-training transfer checks and judge validation, we remove the final user turn from multi-turn conversations and ask the model to generate under the user role. All generation uses \emph{vLLM}\footnote{Our generations can be verified with \href{https://hub.docker.com/layers/verlai/verl/vllm017.latest/}{verlai/verl:vllm017.latest} docker image. We made use of H200 gpus for all experiments.} with modified chat templates that permit user-role continuation. We primarily use temperature~$0$ (deterministic/greedy) generation; additional sampled generations ($T \in \{0.3, 0.7, 1.0\}$) probe whether interaction awareness is present in a model's broader distribution (\S\ref{sec:temperature-results}).

\paragraph{Datasets.} We use $5$ benchmark datasets where assistant behavior is independently verifiable, enabling joint measurement of task accuracy and interaction awareness: GSM8K \citep{cobbe2021training} ($1319$ examples, math reasoning), IFEval \citep{zhou2023ifeval} ($541$ examples) and IFBench \citep{pyatkin2025generalizing} ($300$ examples, instruction following), and GPQA main ($448$ examples) and GPQA diamond ($198$ examples) \citep{rein2024gpqa} (expert QA). On GSM8K we extract the final numeric answer; on GPQA the final letter, with extraction failures counted as incorrect. Held-out datasets are HealthBench \citep{healthbench} ($1470$ examples) and a subset of Coval ($100$ examples) \citep{coval}, where existing human follow-ups provide a natural setup for interaction.

\paragraph{Models.} We evaluate $11$ open-weight models: eight Qwen3.5 models from $0.8$B to $397$B-A$17$B \citep{qwen3.5}, two \texttt{gpt-oss} models ($20$B, $120$B) \citep{openai2025gptoss120bgptoss20bmodel}, and \texttt{GLM-4.7} \citep{zeng2025glm}. The Qwen3.5 sweep provides within-family scaling analysis; cross-model comparison controls for differences in training data and post-training recipes.

\paragraph{Controlled perturbations.}
\label{sec:perturbations}
If the genuine-followup rate measures something real about interaction awareness, it should respond to causal manipulations of the assistant turn. We design two perturbations as positive controls (Figure~\ref{fig:perturbation-examples}). \emph{Truncation} removes $\max(25, \lceil 0.25 \times n_{\text{tokens}} \rceil)$ tokens from the end of the assistant response before generating the user turn; a model that attends to assistant-turn content should produce more genuine follow-ups when the response is visibly incomplete. \emph{Explicit question} appends a question sampled from a pool of generic follow-ups (e.g., ``What do you think?'', ``Any questions?'') to the assistant response, testing sensitivity to a conversational cue rather than an implicit degradation.

\paragraph{Post-training interventions.}
\label{sec:post-training}
To test whether interaction awareness is trainable, we apply the \texttt{CollabLLM} recipe \citep{wu2025collabllm} to \texttt{Qwen3.5-2B}, which optimizes for multi-turn collaboration quality via forward-sampled conversation rewards. Training data comes from $200$ level-5 MATH problems \citep{hendrycks2021measuring}, expanded into $\approx5k$ multi-turn examples ($\approx5$ turns each). We train two variants for $1$ epoch: \textbf{(i)}~SFT with masked loss on assistant turns, and \textbf{(ii)}~online RL with PPO \citep{schulman2017proximal} using multiturn rollouts and conversational rewards. Crucially, neither intervention explicitly trains for user-turn generation; both target assistant response quality only. Any change in follow-up rate therefore reflects an indirect effect of collaboration-aware training. Neither the evaluation benchmarks nor held-out datasets appear in training, so gains reflect transfer.
\begin{figure*}[t]
\centering
\includegraphics[width=\textwidth]{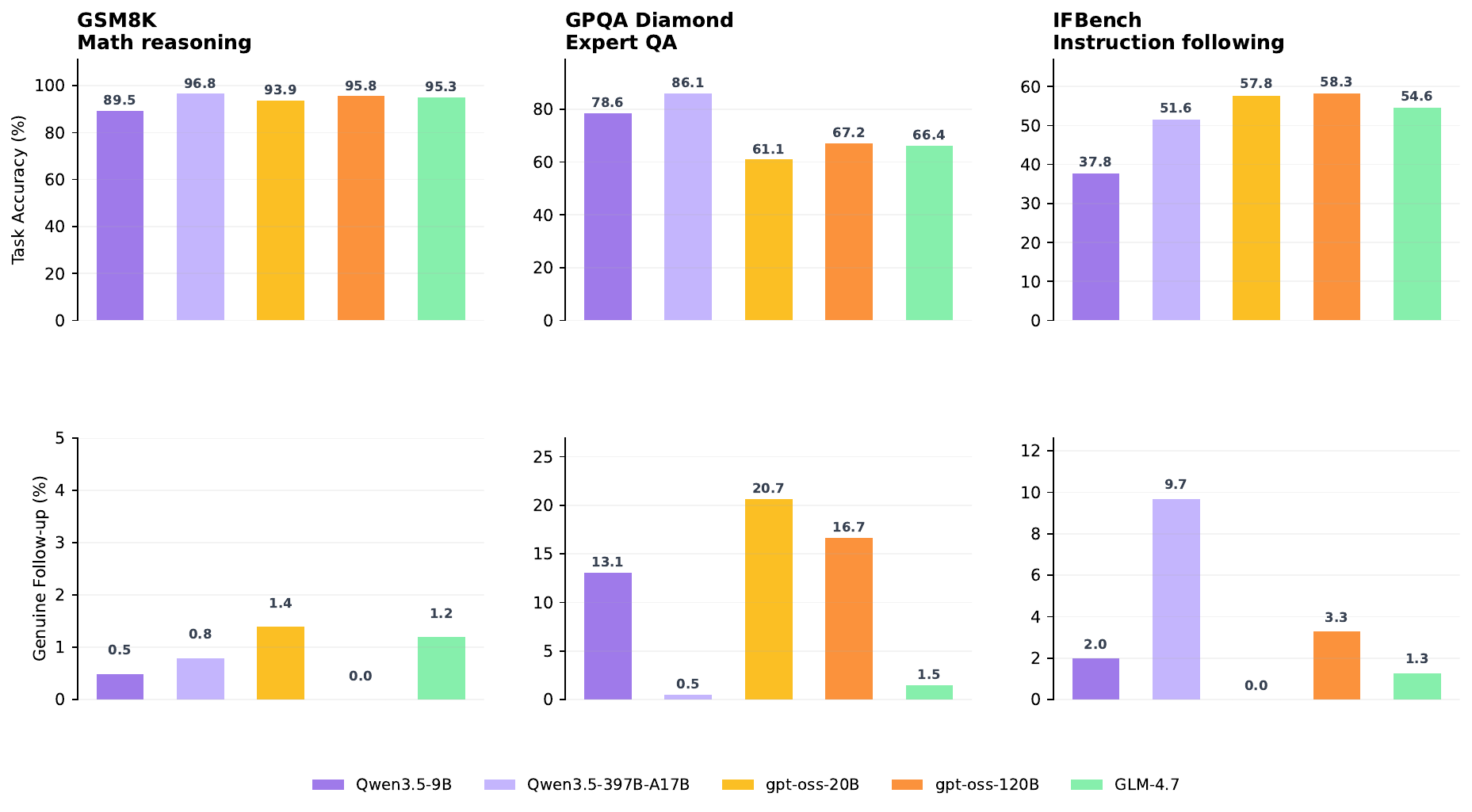}
\caption{\textbf{Cross-family dissociation: task accuracy vs.\ followup rate.} Top row: task accuracy (\%). Bottom row: followup rate (\%). Five representative models on GSM8K, GPQA Diamond, and IFBench. Task accuracy does not predict follow-up quality: \texttt{gpt-oss} models produce the highest follow-up on GPQA despite lower accuracy than Qwen3.5 or \texttt{GLM-4.7}. Full results including IFEval and GPQA Main in Table~\ref{tab:cross-family-full}.}
\label{fig:cross-family}
\end{figure*}

\section{Results}
\label{sec:results}

We present the results of our analysis in five parts: (i) task accuracy does not reflect follow-up quality across model families (\S\ref{sec:cross-family}); (ii) temperature sampling reveals interaction awareness is latent in the generation distribution while appearing absent with deterministic generation (\S\ref{sec:temperature-results}); (iii) a within-family deep dive shows interaction awareness does not scale with model size (\S\ref{sec:scaling}); (iv) controlled perturbations confirm the proposed metric (\S\ref{sec:perturbation-results}); and (v) collaboration-oriented post-training can increase interaction awareness (\S\ref{sec:post-training-results}).

\subsection{Task Accuracy Does Not Predict Follow-Up Quality Across Families}
\label{sec:cross-family}

If interaction awareness were simply a function of general capability, we would expect models with higher task accuracy to also produce better follow-ups. We evaluate representative models from the Qwen3.5 and \texttt{gpt-oss} families alongside \texttt{GLM-4.7} under deterministic generation ($T{=}0$). Figure~\ref{fig:cross-family} presents the results: task accuracy (top row) and followup rate (bottom row) show distinctive behavior and remain largely uncorrelated.

\texttt{gpt-oss} models produce $17$-$21\%$ follow-ups on GPQA Diamond despite lower answer accuracy ($61$-$67\%$) than \texttt{GLM-4.7} ($66.4\%$ accuracy, $1.5\%$ follow-up) or Qwen3.5-397B ($86.1\%$ accuracy, $0.5\%$ follow-up). \texttt{Qwen3.5-397B-A17B} follow up rate is strongest on IFBench ($9.7\%$) but weak on GPQA. Thus, different families develop interaction awareness on different input contexts, and follow-up quality is not a monotone function of answer quality. This provides deeper evidence for why LLMs struggle when repurposed as user proxies for self-play or multi-agent interaction \citep{shekkizhar2025echoing, zhou2026sim2real}: the disconnect is rooted in absent interaction awareness, not insufficient task competence. Appendix~\ref{app:example-generations} gives representative traces for different models: nongenuine cases are typically meta-planning or continuation in assistant persona, while genuine user turns are terse corrections, summary requests, or clarification questions.
\begin{figure*}[t]
\centering
\includegraphics[width=\textwidth]{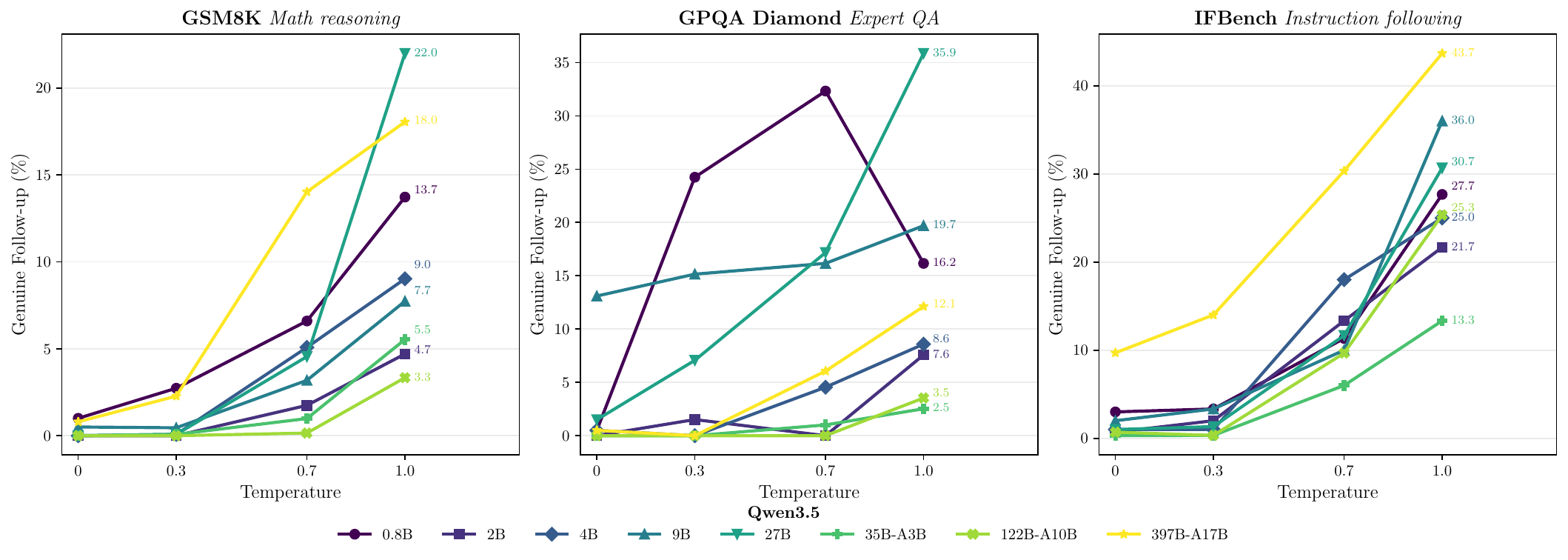}
\caption{\textbf{Qwen3.5 family: interaction awareness across temperature.} Genuine follow-up rate (\%) vs.\ sampling temperature for all eight Qwen3.5 models on three datasets. At $T{=}0$ (greedy), follow-up is near zero for most models despite high task accuracy (Table~\ref{tab:qwen-scaling-full}). Higher temperatures surface latent interaction awareness, but this does not scale monotonically with model size: \texttt{Qwen3.5-9B} and \texttt{Qwen3.5-27B} often match or exceed larger models. MoE variants (\texttt{35B-A3B}, \texttt{122B-A10B}) consistently lag their dense counterparts.}
\label{fig:qwen-scaling}
\end{figure*}

\subsection{Interaction Awareness Is Latent in the Distribution}
\label{sec:temperature-results}

The cross-family results above use deterministic generation ($T{=}0$), which probes only the model's most likely user continuation. A natural question is whether the near-zero follow-up rates reflect a genuine absence of interaction awareness or whether the capability exists in the distribution but is suppressed at the mode. We run a temperature sweep ($T \in \{0, 0.3, 0.7, 1.0\}$) on six representative models (Figure~\ref{fig:overview}, right).

The results reveal a clear separation between model families. Qwen3.5 and GLM show strong temperature scaling: \texttt{Qwen3.5-27B} rises from $0\%$ to $22\%$ on GSM8K, from $1.5\%$ to $35.9\%$ on GPQA Diamond, and from $1\%$ to $30.7\%$ on IFBench. \texttt{GLM-4.7} shows a similar pattern, reaching $15.2\%$ on GSM8K and $35.4\%$ on GPQA Diamond at $T{=}1.0$. In contrast, \texttt{gpt-oss} models demonstrate mixed response to sampling: \texttt{gpt-oss-120b} remains at $0.1\%$ on GSM8K even at $T{=}1.0$, and both \texttt{gpt-oss} models stay below $4\%$ on IFBench across all temperatures. However, \texttt{gpt-oss-20b} reaches $47\%$ at $T{=}1.0$ on GPQA Diamond.

These findings have two implications. First, interaction awareness is latent in the generation distribution of Qwen and GLM models, but current training methods, which optimize for maximum likelihood or low-temperature generation, do not place this capability at the mode. Second, the temperature-resistance of \texttt{gpt-oss} on GSM8K and IFBench indicates a qualitatively different training outcome: these models have not learned interaction-aware continuations for these contexts at any probability. This distinction between latent and genuinely-absent interaction awareness motivates the next section on Qwen3.5 analysis.

\subsection{Within-Family Analysis: Model Size Does Not Predict Interaction Awareness}
\label{sec:scaling}
To disentangle model size from training recipe, we examine the full Qwen3.5 family ($0.8$B to $397$B-A$17$B) under both deterministic and sampled generation. At $T{=}0$, task accuracy scales reliably: GSM8K climbs from $41.6\%$ ($0.8$B) to $96.8\%$ ($397$B-A$17$B), GPQA Diamond from $23.6\%$ to $86.1\%$, and IFBench from $21.6\%$ to $51.6\%$\footnote{Accuracy results on GSM8K was computed while we make use of reported performance in GPQA-Diamond and IFBench from \href{https://artificialanalysis.ai/}{artificialanalysis.ai}}. Follow-up rates, however, stay near zero: five of eight models produce $0.0\%$ on GSM8K (Table~\ref{tab:qwen-scaling-full}). Results on IFEval and GPQA Main are consistent (Appendix~\ref{app:full-tables}).

The full temperature sweep (Figure~\ref{fig:qwen-scaling}) reveals that this non-scaling persists even when interaction awareness is surfaced through sampling. On IFBench at $T{=}1.0$, the largest model (\texttt{397B-A17B}) reaches $43.7\%$, but the $9$B model is close behind at $36\%$, and even the $0.8$B model reaches $27.7\%$. The MoE models (\texttt{35B-A3B}, \texttt{122B-A10B}) consistently lag their dense counterparts. On GPQA Diamond, the $27$B model dominates at $35.9\%$, while the $0.8$B model shows a non-monotonic pattern, peaking at $T{=}0.7$ ($32.3\%$).

Figure~\ref{fig:scaling-contrast} illustrates the qualitative difference: on the same GPQA chemistry question, both \texttt{Qwen3.5-9B} and \texttt{Qwen3.5-27B} answer correctly, but \texttt{Qwen3.5-9B} generates a user turn that critically engages with the assistant's reasoning, while \texttt{Qwen3.5-27B} restates the original prompt verbatim. The mid-range models ($27$B, $35$B-A$3$B, $122$B-A$10$B) are near-uniformly zero at $T{=}0$, dominated by \texttt{previous\_turn\_restate}. Interaction awareness is thus non-monotonic and dataset-dependent, even within a single model family. This has a direct consequence for self-play and multi-agent pipelines: simply using a larger model does not yield a better partner model, and the weights that solve a task do not, as measured by our probe, encode how a user would react to the solution.

\begin{figure*}[t]
\begin{minipage}[t]{0.48\textwidth}
\begin{myboxblue}
{\textbf{Qwen3.5-9B} \hfill \textit{genuine follow-up}}
\begin{customsmall}
\begin{verbatim}
[Assistant] ... The Corey-Chaykovsky reagent
reacts with ketones to form epoxides ...
The final product has 11 carbon atoms.
Answer: D

[User turn]
In the previous turn, you identified the
reagent as dimethylsulfoxonium methylide
(Corey-Chaykovsky reagent) and concluded
that it adds a methylene group to the
ketone, resulting in an epoxide with 11
carbons. However, the reaction conditions
specified "in DMSO at elevated temperature".
Let's re-evaluate the reagent name ...
\end{verbatim}
\end{customsmall}
\end{myboxblue}
\end{minipage}
\hfill
\begin{minipage}[t]{0.48\textwidth}
\begin{mybox}
{\textbf{Qwen3.5-27B} \hfill \textit{prompt restatement}}
\begin{customsmall}
\begin{verbatim}
[Assistant] ... The reagent contributes 1
carbon atom (the methylene group) ...
Total carbons in Product 3: 10 + 1 = 11.
Answer: D

[User turn]
Answer the following multiple-choice
question. Keep the explanation concise.
The last line of your response must be
exactly of the form Answer: <letter>
where <letter> is A, B, C, or D.

Question: trans-cinnamaldehyde was treated
with methylmagnesium bromide ...
\end{verbatim}
\end{customsmall}
\end{mybox}
\end{minipage}
\vspace{-.2cm}
\caption{\textbf{Same question, same correct answer, different interaction awareness.} Both models correctly answer a GPQA chemistry question (Answer: D). \emph{Left:}~\texttt{Qwen3.5-9B} generates a user turn that critically engages with the assistant's reasoning about the Corey-Chaykovsky reagent and reaction conditions. \emph{Right:}~\texttt{Qwen3.5-27B}, a $3\times$ larger model, restates the original prompt verbatim, an inability to generate realistic user turn after an assistant response.}
\label{fig:scaling-contrast}
\end{figure*}

\subsection{Controlled Perturbations Confirm the Metric}
\label{sec:perturbation-results}

Both perturbations move the metric in the expected direction for models that attend to assistant-turn content, confirming that the genuine-followup rate measures something real about interaction awareness.

\paragraph{Truncation.} Table~\ref{tab:perturbations}~(Left) shows the effect of truncating assistant responses. On GSM8K, where unperturbed follow-up is near zero, truncation sharply increases genuine follow-up for \texttt{GLM-4.7} ($1.0\% \to 55.0\%$) and \texttt{gpt-oss-120b} ($0.0\% \to 24.2\%$). The effect extends to GPQA Diamond: \texttt{GLM-4.7} rises from $2.0\%$ to $39.4\%$, and \texttt{gpt-oss-120b} from $20.7\%$ to $65.7\%$. Qwen models are notably less sensitive: \texttt{Qwen3.5-27B} shows no change on GSM8K and remains flat on GPQA Diamond. This family-level asymmetry suggests qualitatively different failure modes: Qwen defaults to prompt restatement regardless of assistant content, while \texttt{gpt-oss} and GLM attend to the assistant turn even when their continuation is not a follow-up.

\begin{table}[t]
\centering
\scriptsize
\begin{minipage}[t]{0.48\textwidth}
\centering
\begin{tabular}{lrrrr}
\toprule
& \multicolumn{2}{c}{GSM8K} & \multicolumn{2}{c}{GPQA Dia.} \\
\cmidrule(lr){2-3} \cmidrule(lr){4-5}
Model & Full & Trunc. & Full & Trunc. \\
\midrule
Qwen3.5-27B & 0.0 & 0.0 & 1.0 & 1.0 \\
Qwen3.5-397B & 0.8 & 4.9 & 2.0 & 2.0 \\
gpt-oss-20b & 1.2 & 5.8 & 1.0 & 18.7 \\
gpt-oss-120b & 0.0 & 24.2 & 20.7 & 65.7 \\
GLM-4.7 & 1.0 & 55.0 & 2.0 & 39.4 \\
\bottomrule
\end{tabular}
\end{minipage}
\hfill
\begin{minipage}[t]{0.48\textwidth}
\centering
\begin{tabular}{lrrr}
\toprule
& \multicolumn{3}{c}{IFBench} \\
\cmidrule(lr){2-4}
Model & Base & +Question & \% Changed \\
\midrule
Qwen3.5-27B & 1.0 & 0.0 & 6.3 \\
Qwen3.5-397B & 7.7 & 4.7 & 40.3 \\
gpt-oss-20b & 1.0 & 6.0 & 97.3 \\
gpt-oss-120b & 1.3 & 25.7 & 99.0 \\
GLM-4.7 & 4.7 & 5.0 & 99.0 \\
\bottomrule
\end{tabular}
\end{minipage}
\caption{\textbf{Controlled perturbations confirm the metric.} Genuine-followup rate (\%) under two perturbations. \emph{Left:}~Truncating the assistant response increases follow-up for non-Qwen models, confirming sensitivity to assistant-turn content. \emph{Right:}~Appending a generic question (e.g., ``What do you think?'') to the assistant response. \emph{Base}: unperturbed follow-up rate; \emph{+Question}: follow-up rate after appending the question; \emph{\% Changed}: fraction of user turns whose text differs from the unperturbed generation, measuring whether the model attends to the perturbation at all. \texttt{gpt-oss} models respond most strongly to both perturbations; Qwen models are largely insensitive. Additional results in Appendix~\ref{app:full-tables}.}
\label{tab:perturbations}
\end{table}

\paragraph{Explicit question.} Table~\ref{tab:perturbations}~(Right) shows the complementary perturbation on the instruction-following datasets. \texttt{gpt-oss-120b} shows the strongest response: genuine follow-up jumps from $1.3\%$ to $25.7\%$ on IFBench, with $99.0\%$ of user turns changing text. \texttt{gpt-oss-20b} follows at smaller magnitude ($1.0\% \to 6.0\%$). \texttt{GLM-4.7} changes $99\%$ of its user turns but genuine follow-up barely increases ($4.7\% \to 5.0\%$ on IFBench). Qwen models again show low sensitivity: \texttt{Qwen3.5-27B} changes only $6.3\%$ of turns, and \texttt{Qwen3.5-397B-A17B} \emph{decreases} in genuine follow-up ($7.7\% \to 4.7\%$ on IFBench), suggesting the perturbation disrupts its existing follow-up behavior rather than eliciting new responses.

Across both perturbations, \texttt{gpt-oss} models are the most responsive: they attend to assistant-turn content and convert that attention into genuine follow-ups. \texttt{GLM-4.7} attends strongly (high changed rates) but does not proportionally increase genuine follow-ups. Qwen models largely ignore the perturbation. These family-specific patterns confirm that interaction awareness is latent and can be surfaced by manipulating the assistant turn (full label-level breakdown in Appendix~\ref{app:failure-modes}). Whether a model can function as a credible conversational partner depends not just on its task accuracy but on how its training recipe shaped its sensitivity to the assistant turn.

\subsection{Post-Training Can Move Interaction Awareness}
\label{sec:post-training-results}

The preceding results establish that interaction awareness is largely absent from current models under default generation. A natural question is whether it can be moved through training. We apply collaboration-oriented post-training to \texttt{Qwen3.5-2B} (\S\ref{sec:post-training}) and find that both SFT and RL raise the genuine-followup rate (Table~\ref{tab:post-training}), despite no overlap between the training data and the evaluation datasets in our results.

\begin{table}[t]
\centering
\scriptsize
\begin{tabular}{lrrrrrr}
\toprule
& GSM8K & \multicolumn{2}{c}{Genuine Follow-up (\%)} & \multicolumn{2}{c}{Held-out} \\
\cmidrule(lr){3-4} \cmidrule(lr){5-6}
Model & Acc. & IFBench & GPQA-Diamond & HealthBench & Coval \\
\midrule
Base & 62.9 & 1.0 & 2.0 & 36.7 & 19.4 \\
SFT & 40.3 & 48.0 & 46.0 & 54.4 & 45.2 \\
RL & 67.4 & 10.0 & 9.1 & 46.5 & 29.0 \\
\bottomrule
\end{tabular}
\caption{Post-training on \texttt{Qwen3.5-2B}. GSM8K accuracy for reference. Followup rates on IFBench, GPQA Diamond and held-out conversational datasets (HealthBench, Coval).}
\label{tab:post-training}
\end{table}

Supervised finetuning~(SFT) produces large gains: follow-up rises from $1.0$-$2.0\%$ to $46$-$48\%$ on verifiable benchmarks and from $36.7\%$ to $54.4\%$ on held-out HealthBench, at the cost of GSM8K accuracy ($62.9\% \to 40.3\%$), likely due to forgetting/overfitting to $\approx 5$k multi-turn examples. The reinforcement learning~(RL) variant avoids this tradeoff, preserving task accuracy ($67.4\%$) while giving moderate follow-up gains across all datasets. Recall that neither intervention explicitly trains for user-turn generation (\S\ref{sec:post-training}); both target assistant response quality only, yet follow-up rates respond in interpretable ways on unseen datasets. These preliminary results suggest the follow-up eval metric can serve as an indicator of collaboration quality and post-training design on this dimension.

\section{Discussion and Conclusion}
\label{sec:discussion}

This work formalizes \emph{user-turn generation} as a probe of interaction awareness in LLMs. Across $11$ open-weight models spanning Qwen3.5, gpt-oss families and \texttt{GLM-4.7}, evaluated on $5$ benchmark datasets, we demonstrate that interaction awareness is largely decoupled from task performance. 
Within Qwen3.5, accuracy scales from $41.6\%$ to $96.8\%$ on GSM8K while grounded follow-up rate stays near zero at every scale. This gap is not uniform: \texttt{gpt-oss} shows relatively higher follow up on GPQA, \texttt{Qwen3.5-397B-A17B} on IFBench, and \texttt{Qwen3.5-9B} as a within-family outlier. 
Controlled perturbations and collaboration-oriented post-training both move the metric in interpretable directions, confirming that interaction awareness is latent, measurable, and can be increased through targeted training.
Both SFT and RL on \texttt{Qwen3.5-2B} raise follow-up rates across datasets despite neither explicitly training for user-turn generation.

Our findings carry four implications. First, assistant-only evaluation misses a dimension that matters for interactive deployment. Second, interaction awareness is not a monotone function of model size or task accuracy, with direct consequences for self-play and multi-agent pipelines. Third, the metric responds to both causal perturbation and training interventions, establishing it as a meaningful axis for post-training design. Fourth, the distinct failure modes across models (Appendix~\ref{app:failure-modes}) suggest interaction awareness is shaped by training recipes rather than being a single unified capability. Crucially, because the probe measures the model's \emph{default behavioral output} under the user role, without additional prompting or instruction, the resulting follow-up rates are directly diagnostic of how training data composition and optimization objectives shape the model's weights. Near-zero follow-up rates possibly reflects training artifacts (e.g., single-turn post-training data, chat template conventions); far from being a confound, this is precisely what the metric is designed to surface in models (see Appendix~\ref{app:alternative-explanations} for more discussion).

\textbf{Limitations.} Our evaluation relies on an LLM judge (\texttt{gpt-5.4-mini}), validated against blinded human annotation ($\kappa = 0.726$) and models. Though limited, future studies should explore alternative detection strategies and more thorough analysis with other judge models. Proposed probe measures whether a model \emph{can} produce a grounded follow-up, not whether it recovers a held-out human continuation. Our held-out evaluations (HealthBench, Coval) partially address this, but both are English-only conversational domains; generalization to multilingual settings, code generation, or longer-horizon multi-turn interactions remains untested. Even in the strongest cells, genuine follow-up remains minority behavior. Downstream utility of the metric, for example, best-of-N assistant response selection, reranking, or self-play training data is left as future work. Our post-training study was preliminary and generalization to other model families remain unexplored. Temperature sweep experiments use a single generation per example; however, since rates are averaged over hundreds of examples per dataset, variance from individual samples is mitigated.

\clearpage
\bibliographystyle{colm2026_conference}
\bibliography{colm2026_conference}
\clearpage

\appendix
\section{Full Results Tables}
\label{app:full-tables}

Tables~\ref{tab:qwen-scaling-full} and~\ref{tab:cross-family-full} report the complete results across all five benchmark datasets, complementing the three-dataset figures in the main text. IFEval and GPQA Main follow the same patterns as IFBench and GPQA Diamond respectively. Table~\ref{tab:qwen-temp-sweep-full} reports the full Qwen3.5 temperature sweep data underlying Figure~\ref{fig:qwen-scaling}.

\begin{table*}[h]
\centering
\small
\begin{tabular}{lrrrrrrr}
\toprule
Model & GSM8K Acc. & GSM8K & IFEval & IFBench & GPQA Main & GPQA Dia.\\
\midrule
Qwen3.5-0.8B & 41.6 & 1.0 & 0.0 & 3.0 & 0.0 & 0.5\\
Qwen3.5-2B & 62.2 & 0.0 & 0.6 & 0.7 & 0.0 & 0.0  \\
Qwen3.5-4B & 86.5 & 0.0 & 1.1 & 1.0 & 0.0 & 0.5 \\
Qwen3.5-9B & 89.5 & 0.5 & 5.4 & 2.0 & 14.7 & 13.1 \\
Qwen3.5-27B & 95.8 & 0.0 & 1.1 & 1.0 & 1.8 & 1.5 \\
Qwen3.5-35B-A3B & 95.8 & 0.0 & 0.7 & 0.3 & 0.0 & 0.0 \\
Qwen3.5-122B-A10B & 95.9 & 0.0 & 0.6 & 0.7 & 0.0 & 0.0 \\
Qwen3.5-397B-A17B & 96.8 & 0.8 & 10.5 & 9.7 & 1.3 & 0.5 \\
\bottomrule
\end{tabular}
\caption{Qwen3.5 scaling: GSM8K answer accuracy (\%) and genuine-followup rate (\%) across all five benchmark datasets.}
\label{tab:qwen-scaling-full}
\end{table*}

\begin{table*}[h]
\centering
\small
\begin{tabular}{lrrrrrrrrrrrrr}
\toprule
& \multicolumn{4}{c}{GSM8K} & \multicolumn{4}{c}{GPQA Diamond} & \multicolumn{4}{c}{IFBench} \\
\cmidrule(lr){2-5} \cmidrule(lr){6-9} \cmidrule(lr){10-13}
Model & $T{=}0$ & $0.3$ & $0.7$ & $1.0$ & $T{=}0$ & $0.3$ & $0.7$ & $1.0$ & $T{=}0$ & $0.3$ & $0.7$ & $1.0$ \\
\midrule
Qwen3.5-0.8B & 1.0 & 2.7 & 6.6 & 13.7 & 0.5 & 24.2 & 32.3 & 16.2 & 3.0 & 3.3 & 11.3 & 27.7 \\
Qwen3.5-2B & 0.0 & 0.0 & 1.7 & 4.7 & 0.0 & 1.5 & 0.0 & 7.6 & 0.7 & 2.0 & 13.3 & 21.7 \\
Qwen3.5-4B & 0.0 & 0.1 & 5.1 & 9.0 & 0.5 & 0.0 & 4.5 & 8.6 & 1.0 & 1.0 & 18.0 & 25.0 \\
Qwen3.5-9B & 0.5 & 0.5 & 3.2 & 7.7 & 13.1 & 15.2 & 16.2 & 19.7 & 2.0 & 3.3 & 10.0 & 36.0 \\
Qwen3.5-27B & 0.0 & 0.1 & 4.5 & 22.0 & 1.5 & 7.1 & 17.2 & 35.9 & 1.0 & 1.3 & 11.7 & 30.7 \\
Qwen3.5-35B-A3B & 0.0 & 0.1 & 1.0 & 5.5 & 0.0 & 0.0 & 1.0 & 2.5 & 0.3 & 0.3 & 6.0 & 13.3 \\
Qwen3.5-122B-A10B & 0.0 & 0.0 & 0.2 & 3.3 & 0.0 & 0.0 & 0.0 & 3.5 & 0.7 & 0.3 & 9.7 & 25.3 \\
Qwen3.5-397B-A17B & 0.8 & 2.3 & 14.0 & 18.0 & 0.5 & 0.0 & 6.1 & 12.1 & 9.7 & 14.0 & 30.3 & 43.7 \\
\bottomrule
\end{tabular}
\caption{Full Qwen3.5 temperature sweep: genuine-followup rate (\%) across all eight models and four temperatures on three datasets.}
\label{tab:qwen-temp-sweep-full}
\end{table*}

\section{Failure Mode Taxonomy}
\label{app:failure-modes}

The LLM judge assigns one of eight primary labels to each generated user turn (\S\ref{sec:metric}). Table~\ref{tab:failure-mode-taxonomy} reports the distribution of these labels across model families on the self-generated benchmark datasets (GSM8K, IFEval, IFBench, GPQA Main, GPQA Diamond). The three families exhibit qualitatively distinct failure profiles.

\begin{table*}[b]
\centering
\small
\begin{tabular}{lrrrrrrr}
\toprule
Model & GSM8K Acc. & GSM8K & IFEval & IFBench & GPQA Main & GPQA Dia. \\
\midrule
Qwen3.5-9B & 89.5 & 0.5 & 5.4 & 2.0 & 14.7 & 13.1  \\
Qwen3.5-27B & 95.8 & 0.0 & 1.1 & 1.0 & 1.8 & 1.5  \\
Qwen3.5-397B-A17B & 96.8 & 0.8 & 10.5 & 9.7 & 1.3 & 0.5 \\
gpt-oss-20b & 93.9 & 1.4 & 3.3 & 0.0 & 14.1 & 20.7 \\
gpt-oss-120b & 95.8 & 0.0 & 3.7 & 3.3 & 17.4 & 16.7 \\
GLM-4.7 & 95.3 & 1.2 & 3.0 & 1.3 & 4.2 & 1.5 \\
\bottomrule
\end{tabular}
\caption{Cross-family genuine-followup rate (\%) on the self-generated benchmark. Includes \texttt{Qwen3.5-9B} as a within-family outlier. GSM8K accuracy shown for reference.}
\label{tab:cross-family-full}
\end{table*}

\begin{table*}[t]
\centering
\small
\begin{tabular}{lrrr}
\toprule
Primary Label & Qwen3.5 (\%) & \texttt{gpt-oss} (\%) & \texttt{GLM-4.7} (\%) \\
\midrule
\texttt{previous\_turn\_restate} & 78.5 & 11.9 & 20.2 \\
\texttt{assistant\_turn\_restate} & 0.0 & 70.2 & 23.0 \\
\texttt{meta\_planning} & 0.4 & 6.8 & 43.0 \\
\texttt{degenerate\_short} & 16.6 & 5.1 & 0.1 \\
\texttt{malformed\_artifact} & 0.8 & 1.0 & 10.8 \\
\texttt{new\_task\_prompt} & 1.4 & 0.8 & 0.9 \\
\texttt{plausible\_followup} & 2.1 & 4.2 & 1.9 \\
\texttt{other} & 0.0 & 0.0 & 0.0 \\
\midrule
$N$ & 5{,}612 & 5{,}612 & 2{,}806 \\
\bottomrule
\end{tabular}
\caption{Failure mode distribution across model families on the self-generated benchmark (all five datasets pooled). Qwen3.5 is dominated by prompt restatement, \texttt{gpt-oss} by assistant-turn restatement (copying the assistant response under the user role), and \texttt{GLM-4.7} by meta-planning (continuing to plan as the assistant despite the user-role header). Each family's dominant failure mode accounts for $43$--$78\%$ of all generations, indicating that interaction awareness failures are systematic and family-specific rather than random.}
\label{tab:failure-mode-taxonomy}
\end{table*}

\textbf{Qwen3.5} defaults to \texttt{previous\_turn\_restate} ($78.5\%$): the model regenerates the original prompt verbatim under the user role, as if starting a new conversation. This pattern is consistent across model sizes, with \texttt{Qwen3.5-27B} and \texttt{Qwen3.5-35B-A3B} reaching nearly $100\%$ restatement on GSM8K. The second most common mode is \texttt{degenerate\_short} ($16.6\%$), particularly for \texttt{Qwen3.5-397B-A17B}, which frequently emits single tokens or empty strings.

\textbf{gpt-oss} shows a fundamentally different failure: \texttt{assistant\_turn\_restate} ($70.2\%$). Rather than restating the prompt, \texttt{gpt-oss} models copy or paraphrase the assistant response under the user role, an identity failure where the model continues in its assistant voice despite the role-header change. This is accompanied by moderate \texttt{meta\_planning} ($6.8\%$), where the model generates planning text (e.g., ``Let me think about how to approach this\ldots'') under the user role.

\textbf{GLM-4.7} is dominated by \texttt{meta\_planning} ($43.0\%$) and \texttt{assistant\_turn\_restate} ($23.0\%$), with a notable \texttt{malformed\_artifact} rate ($10.8\%$). The high meta-planning rate suggests that \texttt{GLM-4.7} attends to the role-header change (unlike Qwen's prompt restatement) but fails to convert that attention into a user-appropriate response, instead producing assistant-style reasoning text.

These distinct failure profiles support the claim in \S\ref{sec:discussion} that interaction awareness is a family of behaviors shaped by training recipes rather than a single unified capability. The perturbation results (\S\ref{sec:perturbation-results}) further confirm this: truncation increases genuine follow-up for \texttt{gpt-oss} and \texttt{GLM} (which attend to the assistant turn but fail to produce follow-ups) but not for Qwen (which ignores the assistant turn entirely).

\section{Representative Generation Examples}
\label{app:example-generations}

We present one dominant failure mode and one genuine follow-up per model family, illustrating the distinct failure profiles described in Appendix~\ref{app:failure-modes}. Each example abridges the original prompt and shows the assistant excerpt and generated user turn.

\paragraph{Qwen3.5-27B: prompt restatement (GSM8K).}
\begin{quote}\small
\textbf{Task (abridged).} ``John arm wrestles 20 people. He beats 80\%. How many did he lose to?''\\
\textbf{Assistant excerpt.} ``Step~1: 80\% of 20 = 16 wins. Step~2: 20 $-$ 16 = 4 losses. Answer:~4''\\
\textbf{Generated user turn.} Verbatim copy of the original prompt, character-for-character.\\
\textbf{Judge.} \texttt{previous\_turn\_restate}; nongenuine.
\end{quote}
This is the modal Qwen failure ($78.5\%$ of all generations). The model ignores the assistant response entirely and restarts the conversation. On GSM8K, \texttt{Qwen3.5-27B} produces $100\%$ restatement.

\paragraph{Qwen3.5-9B: grounded challenge (GPQA Diamond).}
\begin{quote}\small
\textbf{Task (abridged).} Astrophysics question about relativistic travel from the Large Magellanic Cloud.\\
\textbf{Assistant excerpt.} ``\ldots 22 + 81.6 = 103.6 years. Since 103.6 < 150, the astronaut survives \ldots Answer:~C''\\
\textbf{Generated user turn.} ``In the previous question, the distance from the Large Magellanic Cloud to Earth was not explicitly stated in the problem description. How did you determine the distance used in your calculation?''\\
\textbf{Judge.} \texttt{plausible\_followup}; genuine.
\end{quote}
A natural clarification request grounded in a specific gap in the assistant's reasoning. \texttt{Qwen3.5-9B} produces $13.1\%$ genuine follow-ups on GPQA Diamond versus near-zero for most other Qwen sizes.

\paragraph{gpt-oss-120b: internal verification leak (GSM8K).}
\begin{quote}\small
\textbf{Task (abridged).} ``A robe takes 2 bolts of blue fiber and half that much white fiber. How many bolts in total?''\\
\textbf{Assistant excerpt.} ``Blue fiber = 2 bolts. White fiber = half of 2 = 1 bolt. Total = 3. Answer:~3''\\
\textbf{Generated user turn.} \texttt{<|channel|>analysis<|message|>} ``We need to ensure final answer line exactly `Answer: <final answer>' with no extra spaces \ldots The solution must be at most 6 short steps, we have 3 steps, fine \ldots''\\
\textbf{Judge.} \texttt{assistant\_turn\_restate}; nongenuine.
\end{quote}
The \texttt{gpt-oss} dominant failure ($70.2\%$): the model emits internal chain-of-thought verification under the user role, complete with a \texttt{<|channel|>analysis} prefix that is clearly model scaffolding, not a human follow-up.

\paragraph{gpt-oss-120b: grounded continuation (GPQA Diamond).}
\begin{quote}\small
\textbf{Task (abridged).} Chemistry question about optically active compounds.\\
\textbf{Assistant excerpt.} ``We need to count chiral compounds. Evaluate each: (Z)-1-chloro-2-methylbut-1-ene \ldots''\\
\textbf{Generated user turn.} \texttt{<|channel|>analysis<|message|>} ``We need to decide if (1s,3s,5s)-cyclohexane-1,3,5-triol is chiral. Let's draw chair conformation \ldots The stereochemistry S at each carbon defines orientation \ldots''\\
\textbf{Judge.} \texttt{plausible\_followup}; genuine.
\end{quote}
Despite retaining the \texttt{<|channel|>analysis} framing, this example drills into a specific sub-problem that the assistant left partially unresolved, a genuine continuation rather than mere verification.

\paragraph{GLM-4.7: planning as user (IFEval).}
\begin{quote}\small
\textbf{Task (abridged).} ``Write a resume for a fresh high school graduate with at least 12 placeholders in square brackets.''\\
\textbf{Assistant excerpt.} ``[Name] [Address] [Phone Number] [Email Address] \textbf{OBJECTIVE} Motivated and reliable high school graduate \ldots''\\
\textbf{Generated user turn.} ``Here's a thinking process that leads to the suggested resume: 1.~\textbf{Analyze the Request:} Target Audience: Fresh high school graduate. Constraint: Must include at least 12 placeholders \ldots 2.~\textbf{Determine the Structure} \ldots''\\
\textbf{Judge.} \texttt{meta\_planning}; nongenuine.
\end{quote}
GLM's dominant failure ($43\%$): the model generates assistant-style reasoning under the user role, beginning with ``Here's a thinking process.'' It attends to the role-header change but fails to adopt a user perspective.

\paragraph{GLM-4.7: natural revision request (IFEval).}
\begin{quote}\small
\textbf{Task (abridged).} ``Write two different cover letters for a software engineer role at Google---a concise version and a long version.''\\
\textbf{Assistant excerpt.} ``Of course! Here are two cover letters tailored for a Software Engineer role at Google: a concise version and a more detailed, long version \ldots''\\
\textbf{Generated user turn.} ``The concise version is too long. Please shorten it further.''\\
\textbf{Judge.} \texttt{plausible\_followup}; genuine.
\end{quote}
A terse, grounded critique: the user evaluated the output and requested a specific revision. GLM produces such genuine follow-ups in roughly $2$--$3\%$ of cases.

\medskip
These examples illustrate the family-specific failure profiles from Table~\ref{tab:failure-mode-taxonomy}: Qwen restates the prompt, \texttt{gpt-oss} leaks internal verification, and GLM emits planning text. Genuine follow-ups, when they occur, are grounded in specific content of the assistant response, such as, clarification requests, factual challenges, or revision instructions.

\section{LLM Judge Validation and Human Annotation}
\label{app:judge-validation}

To validate the reliability of our LLM-based evaluation system, we conducted both same-family judge comparisons and a blinded manual review study with human annotators. This section details the judge selection rationale, cross-judge agreement, human annotation protocol, and correlation analysis.

\subsection{Same-Family Judge Validation}

Paper-facing tables in this draft use exact-match preprocessing followed by a \texttt{gpt-5.4-mini} judge. We switched from the earlier \texttt{gpt-5.4-nano-2026-03-17} operational judge because same-family judge validation and two blinded human packets both favored \texttt{mini}. The relevant question here is therefore not cross-family agreement, but how much the older nano results contract under \texttt{mini} and whether \texttt{mini} tracks full \texttt{gpt-5.4}. We report three same-family checks. First, we rerun the entire real-follow-up panel with \texttt{gpt-5.4-mini}. Second, we keep a prevalence-preserving random slice with all Coval examples plus 300 random HealthBench examples per model. Third, we keep a balanced hard-case slice with all Coval examples plus 100 HealthBench examples per model, balanced by the nano judge's genuine-versus-nongenuine decision.

\begin{table}[t]
\centering
\small
\begin{tabular}{lrrrrr}
\toprule
Slice / Judge Pair & $N$ & Gen. Agr. & Gen. $\kappa$ & Label Agr. & Label $\kappa$ \\
\midrule
Full panel / \texttt{nano} vs \texttt{mini} & 7815 & 81.7 & 0.480 & 48.0 & 0.396 \\
Random / \texttt{nano} vs \texttt{mini} & 1965 & 82.5 & 0.477 & 48.6 & 0.406 \\
Balanced / \texttt{nano} vs \texttt{mini} & 965 & 72.8 & 0.373 & 47.0 & 0.379 \\
Balanced / \texttt{nano} vs \texttt{gpt-5.4} & 965 & 73.2 & 0.366 & 44.7 & 0.358 \\
Balanced / \texttt{mini} vs \texttt{gpt-5.4} & 965 & 90.9 & 0.722 & 77.3 & 0.721 \\
\bottomrule
\end{tabular}
\caption{Same-family \texttt{GPT-5.4} judge validation. Agreement rates are percentages; $\kappa$ values are unitless. The full-panel row reruns the entire real-follow-up benchmark with \texttt{gpt-5.4-mini}; the random slice is prevalence-preserving, while the balanced slice is a hard-case agreement slice.}
\label{tab:judge-validation}
\end{table}

Table~\ref{tab:judge-validation} gives a more useful signal than a cross-family comparison would. On the full panel, \texttt{gpt-5.4-nano-2026-03-17} and \texttt{gpt-5.4-mini} reach 81.7\% genuine-followup agreement with $\kappa=0.480$, while the overall genuine-followup rate drops from 26.6\% under nano to 18.5\% under mini. That is not enough to call the judges equivalent, but it is enough to support a narrower operational claim: nano is usable for large sweeps on natural data. The contraction is strongly model-dependent, however. Qwen rows remain stable or rise slightly under \texttt{mini}, while \texttt{gpt-oss} and \texttt{GLM-4.7} shrink sharply. The most consequential shift is \texttt{gpt-oss-120b}: it falls from 21.2\% to 4.8\% on HealthBench and from 43.0\% to 16.1\% on Coval. So the appendix should not sell the nano-era raw Coval lead of \texttt{gpt-oss-120b} as a robust ranking result, even though the broader Qwen-versus-non-Qwen family gap survives.

The same table also shows why we should still be careful with the nano judge. The prevalence-preserving random slice lands very close to the full rerun signal: 82.5\% genuine-followup agreement with $\kappa=0.477$, plus moderate file-level rate correlation (Pearson $r=0.642$, Spearman $\rho=0.685$). But the balanced hard-case slice is harsher. There, \texttt{gpt-5.4-nano-2026-03-17} agrees with \texttt{gpt-5.4-mini} and \texttt{gpt-5.4} on only about 73\% of genuine-followup decisions, with $\kappa \approx 0.37$ in both cases. The direction of disagreement is systematic there as well: nano is looser on the hard slice, marking 37.9\% of examples as genuine follow-up, compared with 22.4\% for \texttt{mini} and 18.1\% for \texttt{gpt-5.4}. By contrast, \texttt{gpt-5.4-mini} and \texttt{gpt-5.4} agree strongly on the same balanced slice: 90.9\% genuine-followup agreement with $\kappa=0.722$, and 77.3\% primary-label agreement with $\kappa=0.721$. This does not justify claiming that nano is interchangeable with the larger judges on challenging borderline cases. It does justify the switch made in this draft: paper-facing tables use \texttt{mini}, while the older nano runs remain useful for broad trend checks and for constructing disagreement slices.

\subsection{Manual Review Methodology}
\label{app:manual-review}

To validate the reliability of our LLM-based evaluation system beyond judge-to-judge comparison, we conducted a blinded manual review study with human annotators.

\paragraph{Annotation protocol.} We constructed multiple 100-example annotation packets sampled from the full evaluation panel. Each packet was drawn as a random sample from the mix of all datasets used in the paper. We constructed two types of packets: \emph{hard-case} packets built from \texttt{nano} versus \texttt{gpt-5.4} disagreements plus judge-agree genuine and judge-agree nongenuine examples, and \emph{natural-prevalence} packets sampled proportionally from the full \texttt{nano} versus \texttt{mini} panel. Packets were distributed to peer AI researchers with expertise in LLMs; we obtained $5$ completed annotation returns.

\begin{figure}[ht]
\centering
\includegraphics[width=0.55\textwidth]{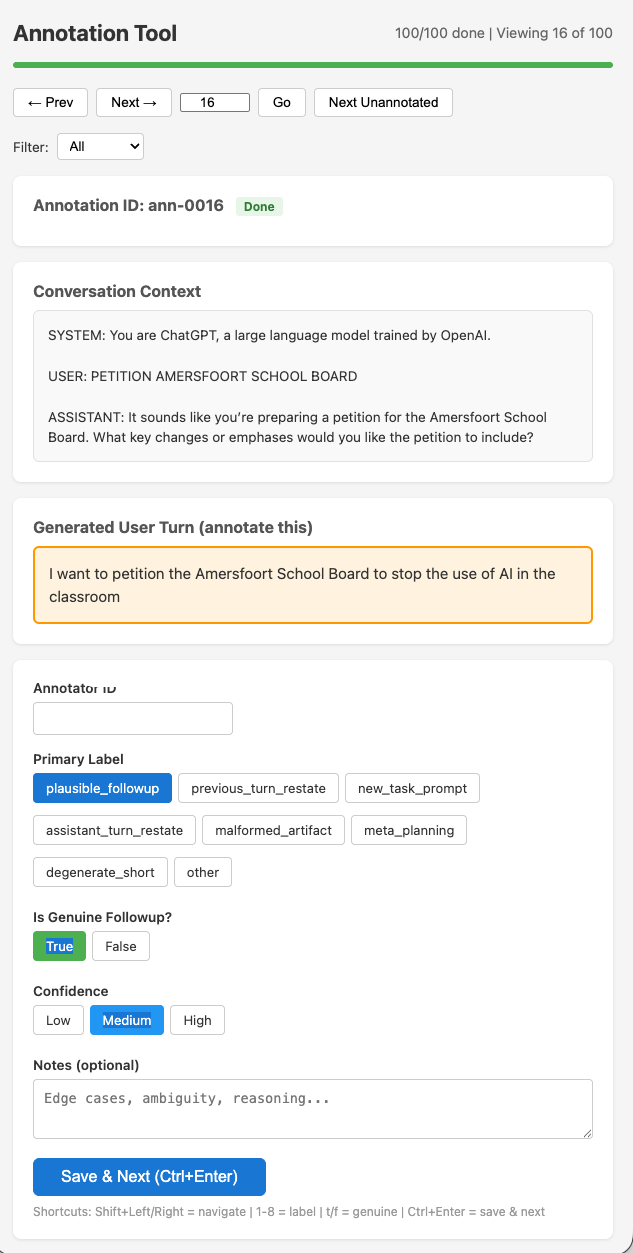}
\caption{\textbf{Manual review interface.} Screenshot of the annotation tool used for LLM judge validation. The interface presents the conversation context and the generated user turn to annotate, along with the primary-label inventory and a binary genuine-followup decision. Model identity, dataset metadata, and judge labels are hidden from annotators.}
\label{fig:annotation-tool}
\end{figure}

Each conversation was presented through a standardized annotation interface (Figure~\ref{fig:annotation-tool}) showing the conversation context and the generated user turn. Annotators did not see the model identity, dataset metadata, reference user turn, or any judge label or rationale, ensuring that human labels were not biased by automated classifications. Annotators were provided with the same evaluation criteria used by the LLM judge (see label definitions below) and assigned a primary label from the same inventory (\texttt{plausible\_followup}, \texttt{previous\_turn\_restate}, \texttt{new\_task\_prompt}, \texttt{assistant\_turn\_restate}, \texttt{malformed\_artifact}, \texttt{meta\_planning}, \texttt{degenerate\_short}, \texttt{other}), a binary genuine-followup decision, and a confidence rating. This design answers whether the benchmark's empirical conclusions survive contact with blinded human judgment, not merely with another LLM judge.

\paragraph{Correlation analysis.} We report human-vs-judge agreement at the binary genuine-followup level, as this is the primary metric used throughout the paper. We did not compute inter-annotator agreement, as our focus was on coverage across packets and correlation with the LLM judge rather than cross-annotator reliability.

On the hard packet, \texttt{gpt-5.4-mini} aligned best with blinded human annotators: 87.0\% genuine-followup agreement with $\kappa=0.688$, versus 67.0\% / 0.255 for \texttt{nano} and 77.0\% / 0.481 for full \texttt{gpt-5.4}. On the natural-prevalence packet, \texttt{mini} again beats \texttt{nano}: 94.0\% genuine-followup agreement with $\kappa=0.766$, versus 79.0\% / 0.354. Pooled across both blinded packets, \texttt{mini} reaches 90.5\% genuine-followup agreement with $\kappa=0.726$, versus 73.0\% / 0.311 for \texttt{nano}. Primary-label agreement remains materially lower (58.5\% with $\kappa=0.481$ for pooled \texttt{mini}), so the human evidence supports the binary judge choice rather than full human equivalence of the entire label inventory.
\section{Related Work}
\label{app:related-work}

\paragraph{Theory of mind in LLMs.}
Evaluating whether LLMs can attribute beliefs and intentions to others has produced a consistent finding: models pass explicit belief-attribution probes but fail when the same reasoning must be expressed behaviorally \citep{sap2022neural, kim-etal-2023-fantom, ma-etal-2023-towards-holistic}. \citet{gu2024simpletom} make this precise: frontier models achieve $>95\%$ on mental-state inference yet drop below $50\%$ on behavior prediction and below $20\%$ on judgment. This illustrates ``knowing versus applying'' dissociation. Our accuracy vs followup gap is structurally analogous, but where SimpleToM probes reasoning about \emph{another agent's} mental state (third-person inference), we probe whether a model encodes the consequence of \emph{its own} response (first-person behavioral readout). Both converge on the same conclusion: solving the immediate task does not guarantee modeling what comes next.

\paragraph{User simulation.}
User simulation has a long history in dialogue systems \citep{schatzmann2007agenda, ShiQWY19}, and same-model self-play has been used for data generation and evaluation \citep{ghandeharioun2019approximating, xu2023baize}. Recent work builds dedicated user proxies: \citet{naous2026flipping} train user models by flipping dialogue turns, \citet{dou2025simulatorarena} benchmark simulator reliability, \citet{hathidara2026mirrorbench} evaluate human-likeness, and \citet{zhou2026sim2real} show that current simulators are overly cooperative and inflate agent scores. Our goal is different: we do not build a reusable user proxy but use user-turn generation as a \emph{probe} of the assistant model's own weights. A high-quality external simulator does not imply that the assistant itself possesses interaction awareness. The failure modes we catalog (prompt restatement, meta-planning, identity leakage) characterize the same behaviors \citet{zhou2026sim2real} observe when assistant models are repurposed as user proxies without dedicated training.

\paragraph{Multi-turn optimization and interaction-aware training.}
Standard post-training optimizes the current response via single-turn preferences or rewards \citep{ouyang2022training, rafailov2023direct}. Recent work extends the training signal beyond the current turn: \citet{wu2025collabllm} optimize for future collaboration quality via forward-sampled multiturn rollouts, \citet{buening2026aligning} use post-action user states as supervision, ArCHer \citep{archer} applies hierarchical multi-turn RL, and \citet{wang2026openclaw} frame every agent interaction as producing a ``next-state signal'' that encodes evaluative and directive information for online learning. In multi-agent RL, \citet{monwilliams2025partner} show that partner models emerge in recurrent agents but only when the environment provides influence over partner behavior, a conditional emergence paralleled by our perturbation results (\S\ref{sec:perturbation-results}), where interaction awareness surfaces only under specific contextual cues. These methods share a premise that the user's reaction carries training signal. Our work provides an evaluation lens for the capability they target: the genuine-followup metric measures the next-state signal that \citet{wang2026openclaw} propose to harness and the collaboration quality that \citet{wu2025collabllm} optimize for, but from an evaluation point of view. Our post-training experiments (\S\ref{sec:post-training-results}) validate this connection, confirming that multiturn collaboration training induces measurable interaction awareness as a side effect.

\paragraph{Interactive benchmarks.}
$\tau^2$-Bench \citep{barres2025tau2}, and CRMArena \citep{huang2025crmarena-pro} evaluate agents in tool-mediated settings with simulated users and task-level rewards, embedding consequence modeling in richer environments. Proactive dialogue work \citep{proactive_survey, star_gate, interactive_text_gen} similarly argues that agents must model the interaction trajectory. Our probe is narrower and this simplicity enables attribution of differences to model weights and training rather than environment confounds. A model that fails our probe lacks the interaction awareness needed for the richer settings these benchmarks target.

\section{Continued Discussion}
\label{app:alternative-explanations}

A natural concern is whether the near-zero follow-up rates on benchmark datasets reflect an absence of interaction awareness or are better explained by artifacts of the training distribution and chat template design. We consider three alternative explanations and discuss the evidence for and against each.

\paragraph{Single-turn training data.}
Models post-trained predominantly on single-turn (query, response) pairs will have learned that the most likely token sequence under the user role is a new query, not a follow-up to the preceding response. Under this explanation, the low follow-up rate is a distributional artifact rather than evidence about what the model's representations encode. We view this as complementary to our framing: if the training data did not include multi-turn continuations, then the model's training did not instill interaction awareness, which is precisely the gap we identify. The post-training results (\S\ref{sec:post-training-results}) directly test this: collaboration-oriented training on multi-turn data raises follow-up rates, confirming that the training distribution is a causal factor.

\paragraph{Chat template conventions.}
The \texttt{<|im\_end|>} token after the assistant turn may function as a strong ``conversation complete'' signal in some training pipelines, biasing the model toward starting a new exchange rather than continuing the current one. If this were the sole explanation, we would expect all models using the same chat template to show similar follow-up rates. However, within the Qwen3.5 family (which shares a template), follow-up rates vary from $0.0\%$ to $13.1\%$ across model sizes and datasets. Cross-family, \texttt{gpt-oss} models produce substantially higher follow-up than Qwen models of comparable or larger size on GPQA, despite all models using role-delimited chat formats. The perturbation results further argue against a purely template-level explanation: truncation and explicit-question perturbations change \texttt{gpt-oss} and \texttt{GLM} follow-up rates substantially while leaving Qwen largely unaffected (\S\ref{sec:perturbation-results}).

\paragraph{Role-header semantics.}
Different model families may have been trained with different conventions for what follows a user-role header (a new independent query versus a continuation of the same conversation). We do not have access to the post-training data for any of the evaluated models, so we cannot rule this out entirely. However, we control for it indirectly through the held-out evaluation setting (\S\ref{sec:experimental-setup}), where the model generates a user turn given a real multi-turn conversation context. On HealthBench and Coval, follow-up rates are substantially higher ($19$--$48\%$) than on benchmark datasets, suggesting that when the conversational context is richer, even models with low benchmark follow-up rates can produce grounded continuations. This is consistent with interaction awareness being latent but context-dependent, rather than absent.

\paragraph{Behavioral versus representational claims.}
We emphasize that our probe measures the model's \emph{behavioral} output under the user role, not its internal representations. A model could in principle encode rich conversational structure internally while still producing restatement as the modal continuation, if its training objective did not incentivize expressing that structure under the user role. Representation-level analyses (e.g., probing internal activations for next-turn content) would complement the behavioral probe we propose. Our claims are therefore about what the model \emph{produces}, conditioned on the experimental setup, rather than about the totality of what the model \emph{knows}. We view this as a feature of the probe rather than a limitation: deployment-relevant interaction awareness must ultimately manifest in generated text, not only in latent representations.

\section{LLM Usage}

For this paper, we made use of LLM tools for three specific purposes: (i) writing code, in particular, for overseeing GPU runs and plotting results from jsonl and csv logs, (ii) review and polish writing, such as, making content concise, checking for statements and claims that were not backed by results in the paper, and (iii) searching for references related to a particular topic. In all scenarios, the responses generated by the LLM were verified for accuracy by the authors, to the best of their knowledge.

\end{document}